\documentclass[conference]{IEEEtran}
\IEEEoverridecommandlockouts

\usepackage{cite}
\usepackage{amsmath,amssymb,amsfonts}
\usepackage{algorithmic}
\usepackage{graphicx}
\usepackage{textcomp}
\usepackage{xcolor}
\usepackage{booktabs}
\usepackage{multirow}
\usepackage{threeparttable}
\usepackage{stfloats}
\usepackage{etoolbox}
\usepackage[hidelinks]{hyperref}

\def\BibTeX{{\rm B\kern-.05em{\sc i\kern-.025em b}\kern-.08em
    T\kern-.1667em\lower.7ex\hbox{E}\kern-.125emX}}

\newcommand{\figref}[1]{Fig.~\ref{#1}}
\newcommand{\tabref}[1]{Table~\ref{#1}}
\newcommand{\eqnref}[1]{Eq.~\eqref{#1}}
\newcommand{\bestperf}{\textcolor{red}{\tiny\,$\blacktriangle$}}
\newcommand{\betterperf}{\textcolor{red}{\tiny\,$\blacksquare$}}

\begin{document}

\title{HFL-FlowLLM: Large Language Models for Network Traffic Flow Classification in Heterogeneous Federated Learning}

\author{%
\IEEEauthorblockN{Jiazhuo Tian\IEEEauthorrefmark{2},\, Yachao Yuan\IEEEauthorrefmark{1}\IEEEauthorrefmark{2}}

\IEEEauthorblockA{\IEEEauthorrefmark{2} School of Future Science and Engineering, Soochow University, Suzhou, China}
\IEEEauthorblockA{jztianjztian@stu.suda.edu.cn, chao910904@suda.edu.cn}
\IEEEauthorblockA{\IEEEauthorrefmark{1} Corresponding author: chao910904@suda.edu.cn}
}

\maketitle

\begin{abstract}
In modern communication networks driven by 5G and the Internet of Things (IoT), effective network traffic flow classification is crucial for Quality of Service (QoS) management and security. Traditional centralized machine learning struggles with the distributed data and privacy concerns in these heterogeneous environments, while existing federated learning approaches suffer from high costs and poor generalization. To address these challenges, we propose HFL-FlowLLM, which to our knowledge is the first framework to apply large language models to network traffic flow classification in heterogeneous federated learning. Compared to state-of-the-art heterogeneous federated learning methods for network traffic flow classification, the proposed approach improves the average F1 score by approximately 13\%, demonstrating compelling performance and strong robustness. When compared to existing large language models federated learning frameworks, as the number of clients participating in each training round increases, the proposed method achieves up to a 5\% improvement in average F1 score while reducing the training costs by about 87\%. These findings prove the potential and practical value of HFL-FlowLLM in modern communication networks security.
\end{abstract}


\section{INTRODUCTION}
Network traffic flow classification is of critical importance to network security. The rapid development of the Internet of Things (IoT) and the wide adoption of 5G and Wi-Fi 6 have led to an explosive growth of heterogeneous devices connected through diverse and dynamic wireless channels. This complex network environment poses significant challenges to the classification of traffic flow. Traditional detection methods rely on centralized data collection and processing, which not only incur huge communication overhead but also bring significant privacy risks. When users are reluctant to share data for privacy reasons, this severely limits the performance of the model. In this distributed environment, Federated Learning (FL) has emerged as a promising method. The preliminary work \cite{mun2020internet} has confirmed the feasibility and accuracy of this method. However, in real world complex wireless network environments, two major challenges are commonly encountered: uneven traffic flow distribution and significant device heterogeneity. Consequently, network traffic flow classification methods based on Heterogeneous Federated Learning (HFL) have emerged, becoming a focal point of subsequent research \cite{marfo2022network, liu2024distributed, guo2022feat, zhu2022attention}.

Although existing methods have achieved promising results in network traffic flow classification under HFL, two major limitations still remain. (i) High design and maintenance costs. The pursuit of accuracy and robustness has led to increasingly complex model, imposing significant design costs and making subsequent maintenance equally costly. (ii) Poor generalization. Although many approaches perform well on specific tasks, their effectiveness deteriorates in unseen environments with different traffic flow tasks or attack patterns.

To address these challenges, Large Language Models (LLM) offer a promising approach for network traffic flow classification. Its strengths in pattern recognition, semantic understanding, and reasoning make it naturally suited for network traffic flow classification. Moreover, LLM's extensive knowledge enables the capture of latent commonalities across diverse traffic patterns, showing strong generalization capabilities. Nevertheless, integrating LLM into HFL also introduces a set of new challenges.

    \textbf{Incompatible architecture.} The autoregressive generation paradigm of mainstream LLM (e.g., ChatGLM2 \cite{zeng2022glm}) is inherently misaligned with the requirements of network traffic flow classification. Token-by-token generation often produces invalid or redundant outputs, while sequential dependencies incur significant inference latency and computational overhead. As a result, directly applying such generative model to classification tasks is suboptimal in both efficiency and accuracy.
    
    \textbf{High training costs and unstable performance.} Mainstream LLM with billions of parameters, are impractical for full fine-tuning on resource-limited HFL clients. A 7B model alone requires about 28 GB VRAM, exceeding most edge devices. Moreover, asynchronous iterations and biased updates arising from client heterogeneity further undermine the performance and stability of the model.
    
    \textbf{Inefficient utilization of client side resources.} In HFL scenarios, network traffic flow are distributed across clients with heterogeneous resources. Uniform training configurations may underutilize high performance clients while excluding low performance clients due to hardware constraints, ultimately degrading global model performance.

In this paper, we aim to address the above challenges and better apply LLM to the network traffic flow classification in the HFL scenario. We design and propose an effective and practical framework, HFL-FlowLLM. (i) To tackle the first challenge, we perform model transformation by replacing the original generative head of the LLM with a network head and compressing the model via layer dropping, thereby enabling efficient inference while significantly reducing latency and uncertainty. (ii) To address the second challenge, we partition the model and apply LoRA fine-tuning exclusively to the modules near the output layers, thereby substantially reducing training costs. During global aggregation, we employ a stacking strategy for noise-free aggregation to mitigate performance drift and instability under heterogeneous scenarios. (iii) To solve the last challenge, we conduct adaptive training for each client that dynamically adjusts the local adaptation scale to fully leverage the computational resources of each client, thereby overcoming bottlenecks in global model performance.
    
    \textbf{Contributions.} Our contributions are summarized as follows:
    \begin{itemize}
        \item We propose HFL-FlowLLM, which to our knowledge is the first framework to apply LLM to network traffic flow classification in HFL.
        \item Compared with existing LLM federated learning frameworks, our approach achieves an optimal balance among model stability, model performance, and the associated training costs through model transformation, adaptive training, and noise-free aggregation.
        \item To evaluate the effectiveness of HFL-FlowLLM, we conducted extensive experiments comparing it with multiple key baselines. The results consistently demonstrate HFL-FlowLLM's superior performance and generalization capability.
    \end{itemize}

\section{RELATED WORK}
\subsection{Network Traffic Flow Classification Based on Federated Learning}\label{TFL}
In recent years, FL has gained wide attention in network traffic flow classification. Early works (e.g., FLIC \cite{mun2020internet}, HFL-ANN \cite{marfo2022network}, FL-CNN-traffic \cite{liu2024distributed}) proved feasibility under idealized IID assumptions, but real-world non-IID conditions significantly degrade accuracy. To counteract this, FEAT \cite{guo2022feat} uses Hoeffding-based skew estimation and Thompson sampling to select low-bias clients. Fed-SOINN \cite{zhu2022attention} and IFLforTFC \cite{pekar2024incremental} adopt incremental learning to ensure the performance of the global model. FLoV2T \cite{zeng2025flov2t} and AECNN \cite{wang2024network} mitigate bias through regularized aggregation and pre-training, respectively. Existing studies mainly focus on the impact of non-IID data distributions, with limited attention to hardware resource constraints. Moreover, their experimental settings are often insufficiently comprehensive to fully demonstrate the generalization capabilities of the proposed methods.

\subsection{Existing LLM Federated Learning Frameworks}
In recent years, integrating FL with LLM has become an important research direction. Early work such as FedIT \cite{zhang2024towards} fine-tunes LLM on local client data and aggregates updates to form a global model, yet full fine-tuning remains infeasible for most clients due to high resource demands. To alleviate this, FedOT \cite{kuang2024federatedscope} adopts model compression and partitioned fine-tuning while maintaining performance. Further improving efficiency, FedBiOT \cite{wu2024fedbiot} proposes a two stage optimization strategy: simulator alignment on the server and adapter alignment on the client, reducing the impact of data distribution discrepancies. Although these approaches utilize LoRA \cite{hu2022lora} based fine-tuning to lower costs, they still face challenges in aggregating LoRA modules of different ranks and remain sensitive to aggregation noise \cite{wang2024FLora}. To mitigate these issues, HETLORA \cite{cho2024heterogeneous} introduces zero-padding for heterogeneous aggregation, and FLoRA \cite{wang2024FLora} employs a stacking mechanism against heterogeneity and noise. However, these methods fail to fully leverage the resources of heterogeneous clients, limiting their effectiveness in HFL scenarios.

\section{METHODOLOGY}
\begin{figure*}[!t]
	\centering
	\includegraphics[width=\textwidth]{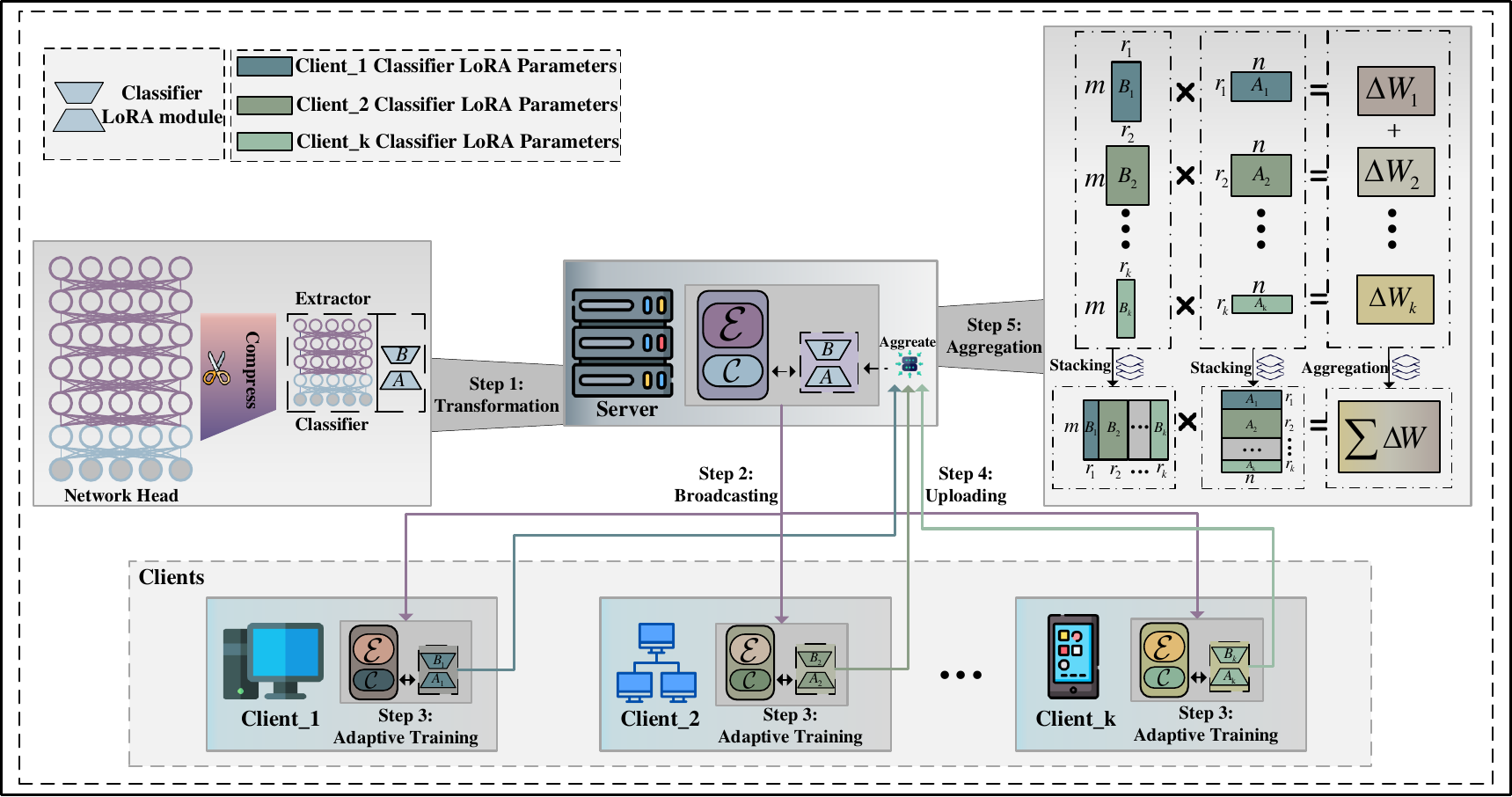}
	\caption{The workflow of HFL-FlowLLM during the HFL training.}
	\label{fig:federatedLearning}
\end{figure*}
As illustrated in \figref{fig:federatedLearning}, the framework consists of five main steps. The following sections provide a detailed explanation of HFL-FlowLLM's working mechanism.
\subsection{Model Transformation}
\begin{figure}[!t]
    \centering
    \includegraphics[width=\linewidth]{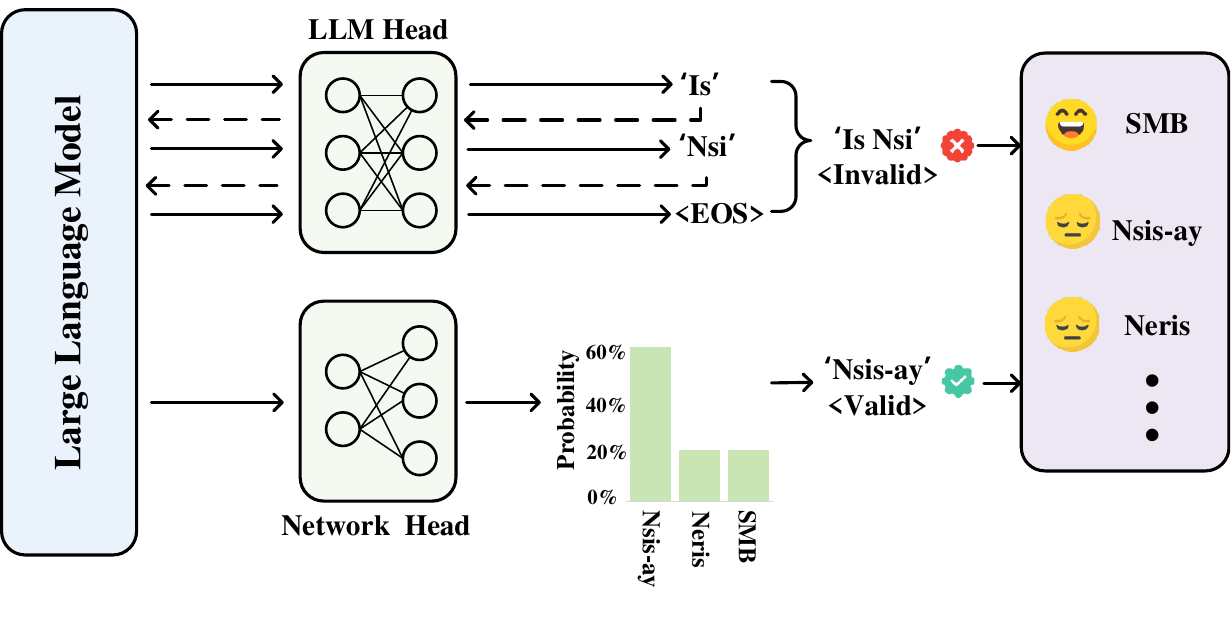}
    \caption{Using malware traffic detection as an example, comparison between the original LLM head and network head.}
    \label{fig:classificationHead}
\end{figure}
\begin{figure}[!t]
    \centering
    \includegraphics[width=\linewidth]{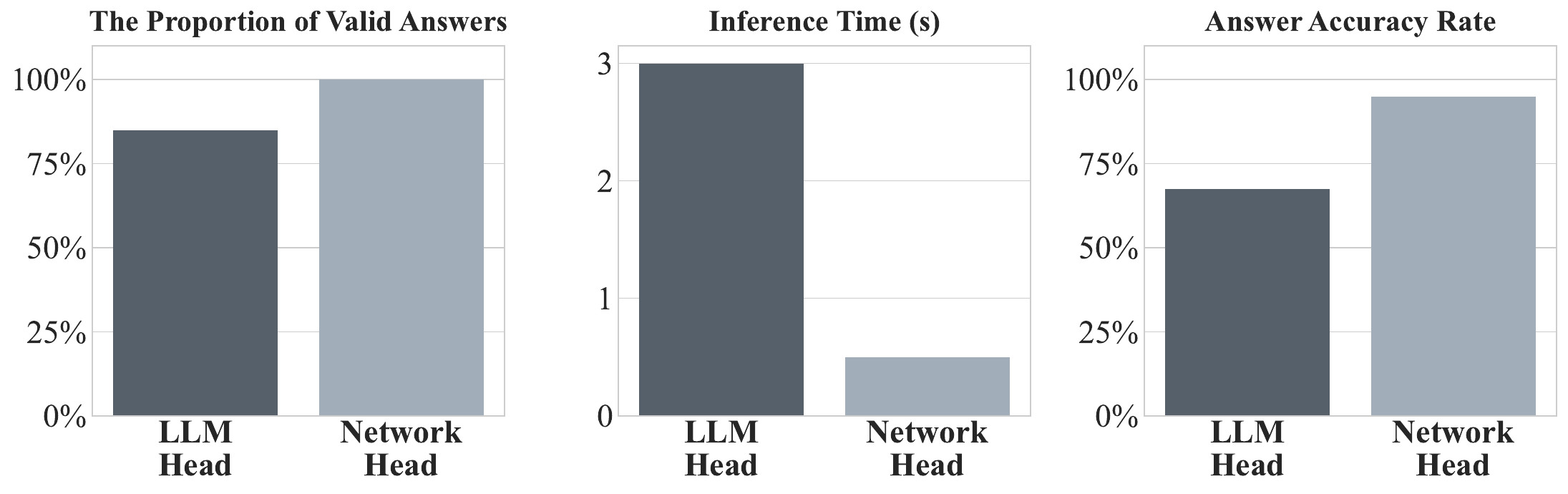}
    \caption{Performance comparison between the LLM head and the network head on malware traffic detection.}
    \label{fig:Head_With_NoHead}
\end{figure}
\textbf{Step 1: Transformation.} Firstly, to fully exploit the potential of LLM, we replace the original output head with a network head specifically designed for network traffic flow classification. Concretely, this network head is implemented as a trainable linear layer that can be flexibly adapted to different network traffic flow classification scenarios. As shown in \figref{fig:classificationHead}, conventional generation model with the LLM head adopts a token-by-token output paradigm. Each output is fed back as input for multiple rounds of inference. Such token-based prediction is inherently uncertain, which may lead to invalid output formats or inaccurate content (e.g., generating non-existent labels such as ``Nsi''). Furthermore, as illustrated in \figref{fig:Head_With_NoHead}, compared with network head, this output mechanism is inferior in terms of output validity, inference time, and prediction accuracy, making it unsuitable for network traffic classification tasks. In contrast, the proposed network head enables the model to produce the final classification result in a single inference step. By selecting the traffic flow category with the highest probability score as output, and since all possible output labels are predefined, the network head strictly constrains predictions to the known set of traffic flow classes, thereby ensuring both the validity and accuracy of the results. 

Secondly, to further reduce the training costs of LLM, we apply layer dropping for model compression. We use $LLM$ to denote the original uncompressed model, $LLM_\text{compress}$ to denote the compressed version, and $index$ represent the index set of the retained layers. The formulation is given as follows:
\begin{equation}
    \label{eq:layer_drop}
    \begin{gathered}
    LLM_\text{compress} \gets LayerExtract(LLM,index)
    \end{gathered}
\end{equation}

Finally, we partition the compressed model $LLM_\text{compress}$ into two components: the part closer to the output layer is denoted as the classifier (represented by $\mathcal{C}$), and the remaining part is denoted as the extractor (represented by $\mathcal{E}$). Formally, we obtain $LLM_\text{compress}$ as a composition of $\mathcal{E}$ and $\mathcal{C}$.
\subsection{Client Update}
\textbf{Step 2: Broadcasting.} After completing Step 1 on the server side, we broadcast $\mathcal{E}$ and $\mathcal{C}$ to all clients. Denote the total number of clients by $K$. The specific computation is defined as follows:
\begin{equation}
    \label{eq:broadcasting}
    \begin{gathered}
    \forall k\in \{1,2\ldots\,K\},\quad  (\mathcal{E},\mathcal{C})\longrightarrow \text{Client}_{k}
    \end{gathered}
\end{equation}
\textbf{Step 3: Adaptive Training.} After receiving $\mathcal{E}$ and $\mathcal{C}$, each client calculates the LoRA rank for $\mathcal{C}$ based on its local resource conditions to make full use of local resources. We construct the three key heterogeneous features of the client $k$: data volume $D_{k}$, data complexity $E_{k}$ (calculated by Shannon entropy), and hardware computing power $C_{k}$ as a resource vector $\mathbf{R}_{k}$ and standardize it. The specific formula is as follows: 
\begin{equation}
    \label{eq:standard}
    \begin{gathered}
    \tilde{\mathbf{R}}_{k} = \frac{\mathbf{R}_{k} - \mathbf{R}_{\min}}{\mathbf{R}_{\max} - \mathbf{R}_{\min}}
    \end{gathered}
\end{equation}
Next, we define a weight vector $\mathbf{w}$, where $\mathbf{w} = [\alpha,\beta,\gamma]^{\top}$, and $\alpha$, $\beta$, and $\gamma$ are the hyperparameters corresponding to the weights of data volume, complexity, and computational capacity, respectively. Moreover, the upper and lower limits of the LoRA rank are preset as $r_{max}$ and $r_{min}$. The LoRA rank for each client can then be calculated as follows:
\begin{equation}
\label{eq:lora-rank-calculation}
\begin{split}
r_k = 2^{\operatorname{round} \left( \log_2 \left( r_{\min} + (r_{\max} - r_{\min}) \cdot \mathbf{w}^\top \tilde{\mathbf{R}}_{k} \right) \right)}
\end{split}
\end{equation}
 In \eqnref{eq:lora-rank-calculation}, $r_k \in [r_{\min},r_{\max}]$, after completing local computations, we initialize two low-rank matrices for the classifier module $\mathcal{C}$ on each client, denoted as $\mathbf{B}_{k}$ and $\mathbf{A}_{k}$. During training, only $\mathbf{B}_{k}$ and $\mathbf{A}_{k}$ are fine-tuned, while the parameters of $\mathcal{E}$ and $\mathcal{C}$ are frozen. Since $\mathcal{E}$ has been extensively pre-trained, it provides sufficient general feature extraction capacity for network traffic classification. In contrast, $\mathcal{C}$, positioned near the output layer, requires fine-tuning to accurately map the deep features from $\mathcal{E}$ to task-specific labels. Therefore, concentrating training on the LoRA module ($\mathbf{B}_{k}$ and $\mathbf{A}_{k}$) of $\mathcal{C}$ is a more effective approach. We use $\theta_{\mathcal{E}}$, $\theta_{\mathcal{C}}$, $\theta_{\mathcal{C}_{\text{LoRA}}}^{(k)}$ represent the parameters of $\mathcal{E}$, $\mathcal{C}$, and the trained LoRA parameters (i.e., $\mathbf{B}_{k}$ and $\mathbf{A}_{k}$). The specific formula is as follows: 
\begin{equation}
\min_{\theta_{\mathcal{C}_{\text{LoRA}}}^{(k)}} \mathcal{L}(\theta_{\mathcal{E}}, \theta_{\mathcal{C}}, \theta_{\mathcal{C}_{\text{LoRA}}}^{(k)}) \quad \text{s.t.} \quad \theta_{\mathcal{E}}, \theta_{\mathcal{C}} \text{ are frozen}
\end{equation}
\subsection{Model Aggregation}
\textbf{Step 4: Uploading.} After completing local training on each client, only $\theta_{\mathcal{C}_{\text{LoRA}}}^{\,(k)}$ is uploaded to the server.

\textbf{Step 5: Aggregation.} After all the client update parameters have been uploaded, we hope to perform the right aggregation as shown in the following formula: 
\begin{equation}
    \label{eq:pk}
    p_{k} = \frac{|D_{k}|}{\sum_{j=1}^{K} |D_{j}|}
\end{equation}
\begin{equation}
\Delta W_{\text{right}} = \sum_{k=1}^{K} p_{k}\theta_{\mathcal{C}_{\text{LoRA}}}^{\,(k)} = \sum_{k=1}^{K} p_{k}\mathbf{B}_{k}\mathbf{A}_{k}
\end{equation}
Where $p_{k}$ represents the weight factor which is jointly determined by the local data volume and the global data volume of all clients. However, applying federated aggregation directly to LoRA leads to the low-rank adaptation matrices $\mathbf{B_{k}}$ and $\mathbf{A_{k}}$ of each client will be weighted respectively:
\begin{equation}
    \mathbf{A}^\ast = \sum_{k=1}^{K}p_{k}\mathbf{A}_{k}, \quad\mathbf{B}^\ast = \sum_{k=1}^{K}p_{k}\mathbf{B}_{k}
\end{equation}
\begin{equation}
    \label{eq:noise}
    \Delta W_{\text{error}} = \mathbf{B}^\ast\mathbf{A}^\ast = \sum_{k=1}^{K}p_{k}^2\mathbf{B}_{k}\mathbf{A}_{k}+\sum _{i\ne j}p_{i}p_{j}\mathbf{B}_{i}\mathbf{A}_{j}
\end{equation}
In \eqnref{eq:noise}, the second item is a noise term, which will reduce the aggregation effect and slow down convergence. To address this issue, we employ a stacking strategy to achieve noise-free aggregation, which is well suited for scenarios with heterogeneous LoRA ranks. The specific formulation is as follows: 
\begin{equation}
    \label{eq:define_stacking_A}
    \begin{gathered}
    \mathbf{A} = p_{1}\mathbf{A}_1 \oplus p_{2}\mathbf{A}_2 \oplus \cdots \oplus p_{k}\mathbf{A}_k, \\ 
    \mathbf{A}_k \in \mathbb{R}^{r_k\times n}, \quad \mathbf{A} \in \mathbb{R}^{(\textstyle \sum_{k=1}^{K}r_{k})\times n}
    \end{gathered}
\end{equation}
\begin{equation}
    \label{eq:define_stacking_B}
    \begin{gathered}
    \mathbf{B} = \mathbf{B}_1 \otimes \mathbf{B}_2  \otimes \cdots \otimes  \mathbf{B}_k, \\ 
    \mathbf{B}_k \in \mathbb{R}^{m\times r_k}, \quad \mathbf{B} \in \mathbb{R}^{m\times (\textstyle \sum_{k=1}^{K}r_{k})}
    \end{gathered}
\end{equation}
We use ``$\oplus$'' to indicate that, for $\mathbf{A}$, each subsequent module is stacked vertically below the preceding one, and ``$\otimes$'' to indicate that, for $\mathbf{B}$, each module is stacked horizontally to the right of the previous one. So we can obtain the $\Delta W_{\text{right}}$ of noise-free aggregation as follows: 
\begin{equation}
    \label{eq:FLora_stacking}
    \begin{split}
    \Delta W_{\text{right}} = (\mathbf{B}_1 \otimes \dots \otimes \mathbf{B}_k)\cdot (p_{1}\mathbf{A}_1 \oplus \dots \oplus p_{k}\mathbf{A}_k)
    \end{split}
\end{equation}
The entire learning process will repeat steps 2 to 5 above until the performance of the global model meets the convergence criterion.
\section{EXPERIMENTAL EVALUATION}
\subsection{Experimental Datasets}
To comprehensively evaluate HFL-FlowLLM, we use five representative public datasets covering key network security scenarios as shown in \tabref{tab:implementation-traffic-dataset}. These datasets capture diverse traffic characteristics providing a robust basis for assessing model performance and generalization.
    \begin{table}[t]
        \centering
        \footnotesize
        \caption{Details of the five traffic flow datasets used in our experiments.}
        \label{tab:implementation-traffic-dataset}
        \begin{tabular}{@{}l c c@{}}
            \toprule
            \textbf{Tasks} & \textbf{Datasets} & \textbf{Labels} \\
            \midrule
            Malware Traffic Detection & USTC TFC 2016 \cite{wang2017malware} & 20 \\
            Botnet Detection & ISCX BOTNET 2014 \cite{beigi2014towards} & 5 \\
            Encrypted VPN Detection & ISCX VPN 2016 \cite{gil2016characterization} & 14 \\
            Tor Behavior Detection & ISCX TOR 2016 \cite{lashkari2017characterization} & 8 \\
            Encrypted App Classification & CSTNET 2023 \cite{lin2022bert} & 20 \\
            \bottomrule
        \end{tabular}
    \end{table}
\subsection{Experimental Environment and Hyperparameter Settings}
All experiments are conducted on Ubuntu 22.04.4 LTS with PyTorch 2.5.1. The hardware platform comprises multiple high-performance GPUs (two NVIDIA GeForce RTX 4090 and one A100) and a server with 512 GB of RAM. ChatGLM2-6B serves as the base LLM. Datasets are split into training, validation, and test sets with an 8:1:1 ratio. To simulate realistic HFL scenarios, training data is distributed across 10 clients via a Dirichlet distribution with concentration factor 0.2, with VRAM allocations (in GB) of [48, 48, 48, 48, 48, 24, 24, 24, 12, 12]. Based on experience, during the compression process, the $index$ are set to the 13 layers close to the input and the 5 layers close to the output, reducing the original 28-layer list to 18 layers. HFL-FlowLLM is evaluated under three participation settings—3, 5, or 8 clients per round—over 10 communication rounds. Each client trains locally for one epoch per round, using an initial learning rate of $3\times 10^{-4}$ and a maximum input length of 1024.

\subsection{Baselines and Evaluation Metrics}
To evaluate HFL-FlowLLM, we compare it against representative benchmarks from two categories. (i) HFL methods for network traffic flow classification. (ii) state-of-the-art LLM federated fine-tuning frameworks.

In the context of HFL-based network traffic flow classification, we select six representative benchmark methods: 
\begin{itemize}
    \item \textbf{Supervised training.} FL-CNN-traffic \cite{liu2024distributed}, FLIC \cite{mun2020internet}, HFL-ANN \cite{marfo2022network}, FEAT \cite{guo2022feat}.
    \item \textbf{Semi-supervised training.} FEDDBN-IDS \cite{nivaashini2024feddbn}, AECNN \cite{wang2024network}.
\end{itemize}
For LLM FL frameworks, we selected FedIT \cite{zhang2024towards}, which employs full model fine-tuning, and FedOT \cite{kuang2024federatedscope}, which uses partial fine-tuning, as baseline methods for comparison.

This study adopts four standard classification metrics: accuracy (ACC), precision (PR), recall (RC), and F1 score (F1) to conduct a comprehensive assessment of the framework.
\subsection{Experimental Results and Analysis}
    \begin{table*}[!t]
          \centering
          \caption{Network traffic flow classification results in HFL.}
          \label{tab:traffic_results_FL}
          \footnotesize
          \setlength{\tabcolsep}{3pt} 
          \begin{tabular}{@{}l ccc ccc ccc ccc ccc@{}}
            \toprule
            \multirow{2}{*}{\textbf{Method}} & \multicolumn{3}{c}{\textbf{USTC TFC 2016}} & \multicolumn{3}{c}{\textbf{ISCX BOTNET 2014}} & \multicolumn{3}{c}{\textbf{CSTNET 2023}} & \multicolumn{3}{c}{\textbf{ISCX VPN 2016}} & \multicolumn{3}{c}{\textbf{ISCX TOR 2016}} \\
            \cmidrule(lr){2-4} \cmidrule(lr){5-7} \cmidrule(lr){8-10} \cmidrule(lr){11-13} \cmidrule(lr){14-16}
            & \textbf{PR} & \textbf{RC} & \textbf{F1} & \textbf{PR} & \textbf{RC} & \textbf{F1} & \textbf{PR} & \textbf{RC} & \textbf{F1} & \textbf{PR} & \textbf{RC} & \textbf{F1} & \textbf{PR} & \textbf{RC} & \textbf{F1} \\
            \midrule
            FL-CNN-traffic \cite{liu2024distributed} & 0.8636 & 0.8327 & 0.7942 & 0.9126 & 0.8231 & 0.8519 & 0.6950 & 0.6731 & 0.6523 & \underline{0.8971} & \underline{0.9090} & \underline{0.8933} & 0.9066 & 0.8081 & 0.7861 \\
            FEDDBN-IDS \cite{nivaashini2024feddbn}    & \underline{0.8889} & \underline{0.8341} & \underline{0.8263} & 0.8913 & 0.8874 & 0.8522 & 0.6691 & 0.5525 & 0.5328 & 0.7814 & 0.7046 & 0.7459 & 0.9295 & 0.9042 & \underline{0.9029} \\
            FLIC \cite{mun2020internet}        & 0.8574 & 0.8821 & 0.8631 & 0.7258 & 0.7361 & 0.6914 & 0.4442 & 0.2091 & 0.2720 & 0.7782 & 0.7370 & 0.7731 & 0.7106 & 0.7266 & 0.6884 \\
            HFL-ANN  \cite{marfo2022network}            & 0.8766 & 0.8287 & 0.8152 & \underline{0.9148} & \underline{0.8926} & \underline{0.8974} & 0.6164 & 0.6092 & 0.5884 & 0.8644 & 0.8503 & 0.8447 & 0.9271 & 0.8913 & 0.9006 \\
            AECNN  \cite{wang2024network}  & 0.8695  & 0.8130 & 0.7995 & 0.7324 & 0.7103 & 0.6928 & 0.6192 & 0.6088 & 0.6237& 0.7829 & 0.7214 & 0.7102 & 0.8021 & 0.7923 & 0.8261 \\
            FEAT \cite{guo2022feat} & 0.8252 & 0.8333 & 0.8217 & \underline{0.8825} & \underline{0.8907} & \underline{0.8888} & \underline{0.7634} & \underline{0.7812} & \underline{0.7723} & 0.8333 & 0.8760 & 0.8072 & \underline{0.8882} & \underline{0.8615} & \underline{0.8486} \\
            \midrule
            \textbf{HFL-FlowLLM} & \textbf{0.9120} & \textbf{0.9050} & \textbf{0.8785} & \textbf{0.9641} & \textbf{0.9640} & \textbf{0.9640} & \textbf{0.9506} & \textbf{0.9430} & \textbf{0.9411} & \textbf{0.9471} & \textbf{0.9454} & \textbf{0.9445} & \textbf{0.9629} & \textbf{0.9624} & \textbf{0.9599} \\
            \bottomrule
          \end{tabular}
    \end{table*}
    \begin{table*}[!ht]
        \centering
        \caption{Comparison between HFL-FlowLLM and other baseline frameworks.}
        \label{tab:fed_performance}
        \footnotesize
        \setlength{\tabcolsep}{5pt}
        \begin{threeparttable}
        \begin{tabular}{@{}llccccccccc@{}}
        \toprule
        \multirow{2}{*}{\textbf{Dataset}} & \multirow{2}{*}{\textbf{Framework}} & \multicolumn{3}{c}{\textbf{3 Clients per Round}} & \multicolumn{3}{c}{\textbf{5 Clients per Round}} & \multicolumn{3}{c}{\textbf{8 Clients per Round}} \\ \cmidrule(lr){3-5} \cmidrule(lr){6-8} \cmidrule(lr){9-11}
         &  &\textbf{PR} &\textbf{RC} &\textbf{F1} &\textbf{PR} &\textbf{RC} &\textbf{F1} &\textbf{PR} &\textbf{RC} &\textbf{F1} \\ 
        \midrule
        \multirow{3}{*}{\shortstack[l]{USTC TFC 2016}}
         & FedIT &0.7384  &0.8182  &0.7631  &0.6834  &0.7094  &0.6463  &0.7054  &0.7925  &0.7307  \\
         & FedOT &0.5851  &0.6430  &0.5746  &0.7059  &0.7283  &0.6786  &0.6653  &0.7011  &0.6410  \\
         & HFL-FlowLLM &\textbf{0.8366}\bestperf  &\textbf{0.8391}\bestperf  &\textbf{0.8342}\bestperf  &\textbf{0.9047}\bestperf  &\textbf{0.8621}\bestperf  &\textbf{0.8476}\bestperf  &\textbf{0.9120}\bestperf  &\textbf{0.9050}\bestperf  &\textbf{0.8785}\bestperf  \\ 
        \midrule
        \multirow{3}{*}{\shortstack[l]{ISCX BOTNET 2014}}
         & FedIT &0.8334  &\textbf{0.8310}  &\textbf{0.8308}  &0.8468  &0.8320  &\textbf{0.8298}  &0.9418  &0.9370  &0.9366  \\
         & FedOT &0.7065  &0.8060  &0.7406  &0.8955  &0.7960  &0.7305  &0.8855  &0.8571  &0.8431  \\
         & HFL-FlowLLM &\textbf{0.8540}\bestperf  &0.8130\betterperf  &0.7914\betterperf  &\textbf{0.8965}\bestperf  &\textbf{0.8332}\bestperf  &0.8084\betterperf  &\textbf{0.9641}\bestperf  &\textbf{0.9640}\bestperf  &\textbf{0.9640}\bestperf \\ 
        \midrule
        \multirow{3}{*}{\shortstack[l]{CSTNET 2023}}
         & FedIT &0.8115  &\textbf{0.8240}  &0.7810  &0.9249  &0.9010  &0.8964  &0.9442  &0.9390  &0.9359  \\
         & FedOT &0.8150  &0.8171  &0.7622  &0.8786  &0.8320  &0.8138  &0.9100  &0.8984  &0.8854  \\
         & HFL-FlowLLM &\textbf{0.8771}\bestperf  &0.7992  &\textbf{0.7957}\bestperf  &\textbf{0.9254}\bestperf  &\textbf{0.9033}\bestperf  &\textbf{0.8988}\bestperf  &\textbf{0.9506}\bestperf  &\textbf{0.9430}\bestperf  &\textbf{0.9411}\bestperf \\
         \midrule
        \multirow{3}{*}{\shortstack[l]{ISCX VPN 2016}}
         & FedIT &0.8570  &0.8620  &0.8273  &0.9289  &0.9230  &0.9219  &0.8959  &0.8742  &0.8672  \\
         & FedOT &0.8473  &0.8351  &0.8018  &0.8918  &0.8700  &0.8591  &0.8561  &0.8172 &0.7808  \\
         & HFL-FlowLLM &\textbf{0.8768}\bestperf  &\textbf{0.9130}\bestperf  &\textbf{0.8885}\bestperf  &\textbf{0.9603}\bestperf  &\textbf{0.9560}\bestperf  &\textbf{0.9561}\bestperf  &\textbf{0.9471}\bestperf  &\textbf{0.9454}\bestperf  &\textbf{0.9445}\bestperf \\
         \midrule
         \multirow{3}{*}{\shortstack[l]{ISCX TOR 2016}}
         & FedIT &\textbf{0.9037}  &\textbf{0.8830}  &0.8616  &0.9190  &0.9130  &0.9115  &0.9190  &0.9132  &0.9115  \\
         & FedOT &0.8786  &0.8630  &0.8592  &0.9338  &0.9270  &0.9263  &0.9160  &0.9123  &0.9100  \\
         & HFL-FlowLLM &0.8756  &0.8760\betterperf  &\textbf{0.8786}\bestperf  &\textbf{0.9469}\bestperf  &\textbf{0.9440}\bestperf  &\textbf{0.9436}\bestperf &\textbf{0.9629}\bestperf  &\textbf{0.9624}\bestperf  &\textbf{0.9599}\bestperf \\
        \bottomrule
        \end{tabular}
        \begin{tablenotes}
            \item[$\dagger$] \footnotesize The \bestperf~symbol indicates that the value is higher than both FedIT and FedOT. The \betterperf~symbol indicates the value is higher than FedOT only.
            \item[$\ddagger$] \footnotesize To fully demonstrate the advantages of our framework, the training of all frameworks adopts the model integrating the network head.
        \end{tablenotes}
        \end{threeparttable}
    \end{table*}
    
    As shown in \tabref{tab:traffic_results_FL}, HFL-FlowLLM consistently outperforms all baselines across tasks, demonstrating superior classification and generalization capabilities. It achieves F1 scores of 0.9640 and 0.9599 on the ISCX BOTNET and ISCX TOR datasets, respectively. On the CSTNET 2023 dataset, our method achieves an F1 of 0.9411, representing a 16.88\% improvement over the best baseline and highlighting its effectiveness on complex encrypted traffic. Even on the challenging malware detection task, it achieves 0.8785, surpassing all baselines. These results confirm the high accuracy and robustness of HFL-FlowLLM in network traffic flow classification.
    \begin{figure}[!t]
        \centering
        \includegraphics[width=\linewidth]{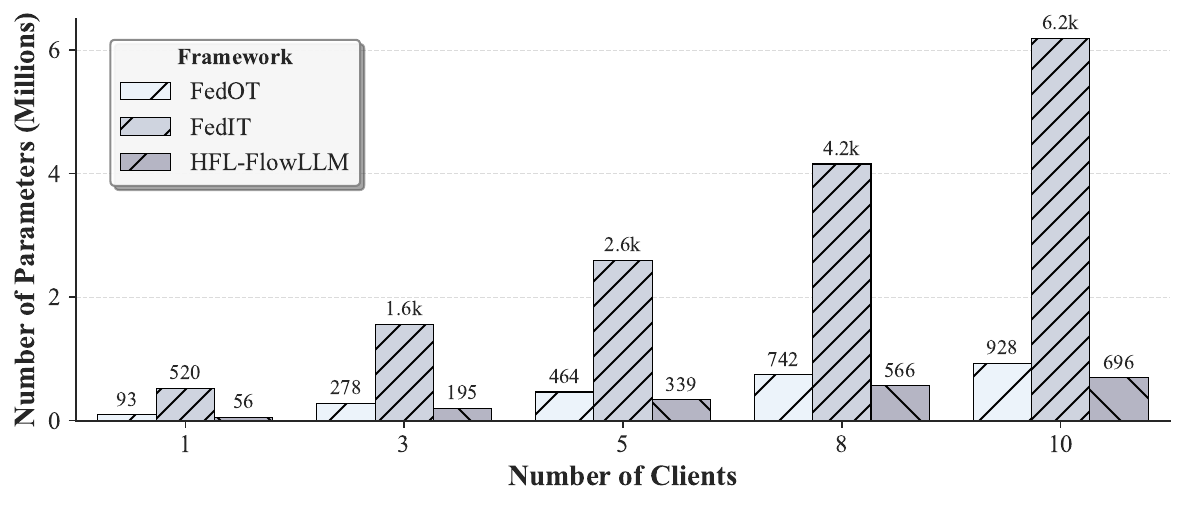}
        \caption{Using malware traffic detection as an example, comparison of training parameter counts across frameworks.}
        \label{fig:federated_learning_params_comparison}
    \end{figure}
    
    As shown in \tabref{tab:fed_performance}, HFL-FlowLLM consistently outperforms baseline frameworks. On USTC-TFC and ISCX-VPN, it achieves superior results across all client settings, while on CSTNET its scalability is evident as the F1 score rises from 0.7957 to 0.9411 when clients increase from 3 to 8. For ISCX-BOTNET and ISCX-TOR, FedIT performs better with fewer clients, but HFL-FlowLLM surpasses it at 8 clients. As illustrated in \figref{fig:federated_learning_params_comparison}, this results from their tuning strategies: FedIT uses full-parameter fine-tuning, which mitigates noise effects with fewer clients and facilitates higher performance, while HFL-FlowLLM adopts noise-free aggregation, ensuring that performance improves steadily as the number of clients increases without introducing noise. Consequently, HFL-FlowLLM demonstrates both low training overhead and stable performance.
    \begin{figure}[!t]
        \centering
        \includegraphics[width=\linewidth]{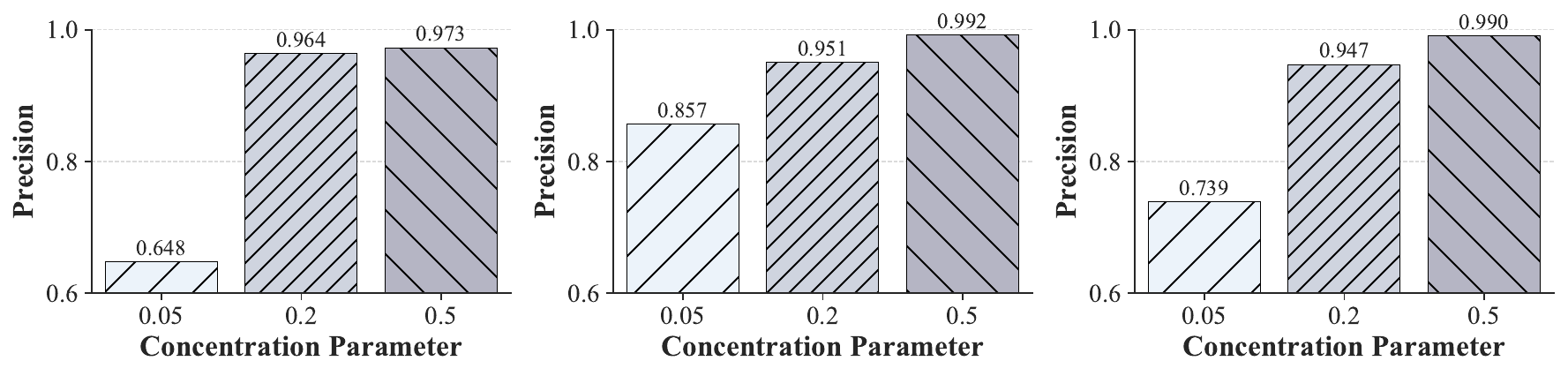}
        \caption{PR of HFL-FlowLLM under different Dirichlet concentrations on ISCX BOTNET 2014, CSTNET 2023, and ISCX VPN 2016.}
        \label{fig:influence_factor_comparison}
    \end{figure}
    
    To evaluate HFL-FlowLLM under varying data distributions, the Dirichlet concentration parameter $\sigma$ was set to $\{0.5, 0.2, 0.05\}$, corresponding to near-IID, moderate non-IID, and highly non-IID partitions, respectively. As shown in \figref{fig:influence_factor_comparison}, under near-IID conditions ($\sigma = 0.5$), HFL-FlowLLM achieves high performance, with PR scores exceeding 0.97 on all datasets, demonstrating strong feature extraction and classification capabilities. As $\sigma$ decreases, performance declines due to distribution skew. Notably, it maintains robustness on CSTNET with a PR score of 0.857 at $\sigma = 0.05$, and exceeds 0.64 on other tasks. These results confirm that HFL-FlowLLM preserves competitive performance and robustness under different challenging non-IID conditions.
    \begin{figure}[!t]
        \centering
        \includegraphics[width=\linewidth]{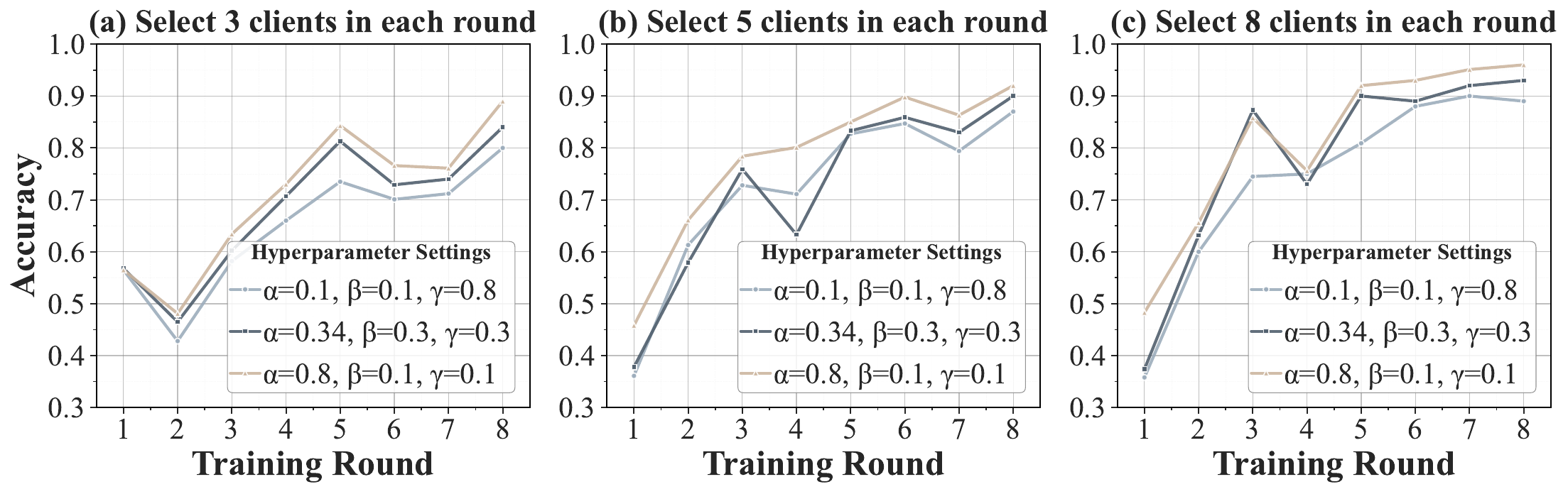}
        \caption{Using the botnet detection as an example, we evaluate the impact of different weighting strategies on classification accuracy.}
        \label{fig:hyperparameter_comparison}
    \end{figure}
    
    In order to better test the adaptive rank mechanism in HFL-FlowLLM, we designed three hyperparameter configurations to evaluate it. As illustrated in \figref{fig:hyperparameter_comparison}, the weight configuration prioritizing data volume ($\alpha = 0.8$, $\beta = 0.1$, $\gamma = 0.1$) achieves optimal performance across nearly all test scenarios, with a consistently higher accuracy curve in most rounds, aligning with the intuition that emphasizing data rich clients enhances feature learning when hardware resources are sufficient. While the balanced strategy ($\alpha = 0.34$, $\beta = 0.33$, $\gamma = 0.33$) offers robust and stable convergence by mitigating extreme metric biases, and the computation focused strategy ($\alpha = 0.1$, $\beta = 0.1$, $\gamma = 0.8$) prioritizes resource constrained yet data critical clients to avoid local optima. The above results indicate that HFL-FlowLLM can fully utilize resources for LLM training in different scenarios.

\section{CONCLUSION}
In this work, we propose a novel framework, HFL-FlowLLM, which is designed to effectively apply LLM to network traffic flow classification in HFL. By designing and optimizing five key steps: transformation, broadcasting, adaptive training, uploading, and aggregation. We successfully address challenges such as architectural incompatibility, high training costs, unstable performance and inefficient utilization of resources. Comprehensive experiments conducted on five public datasets demonstrate that HFL-FlowLLM significantly outperforms existing methods. In summary, HFL-FlowLLM offers a new perspective for traffic classification, providing a powerful and practical solution that paves the way for broader adoption of LLM in distributed network security.
\vspace{12pt}

\bibliographystyle{IEEEtran}
\bibliography{main}

\end{document}